\documentclass[letterpaper, 10 pt, conference]{ieeeconf}  

\IEEEoverridecommandlockouts                              

\usepackage{graphics} 
\usepackage{epsfig} 
\usepackage{amsmath} 
\usepackage{amssymb}  
\usepackage{amsfonts}
\usepackage[noadjust]{cite}
\usepackage{color}
\usepackage{multirow}
\usepackage{array}
\usepackage{caption}
\usepackage{booktabs}
\usepackage{makecell}
\usepackage{graphicx}
\usepackage[caption=false]{subfig}

\title{\LARGE \bf
Simultaneous Localisation and Mapping with Quadric Surfaces
}

\author{Tristan Laidlow and Andrew J. Davison%
\thanks{The authors are with the Dyson Robotics Lab at Imperial College, Imperial College London, UK.}%
\thanks{Research presented in this paper has been supported by Dyson Technology Ltd. Corresponding author: Tristan Laidlow, {\tt\small t.laidlow15@imperial.ac.uk}}%
}

\begin{document}

\maketitle
\thispagestyle{empty}
\pagestyle{empty}

\begin{abstract}

There are many possibilities for how to represent the map in simultaneous localisation and mapping (SLAM).
While sparse, keypoint-based SLAM systems have achieved impressive levels of accuracy and robustness, their maps may not be suitable for many robotic tasks.
Dense SLAM systems are capable of producing dense reconstructions, but can be computationally expensive and, like sparse systems, lack higher-level information about the structure of a scene.
Human-made environments contain a lot of structure, and we seek to take advantage of this by enabling the use of quadric surfaces as features in SLAM systems.
We introduce a minimal representation for quadric surfaces and show how this can be included in a least-squares formulation.
We also show how our representation can be easily extended to include additional constraints on quadrics such as those found in quadrics of revolution.
Finally, we introduce a proof-of-concept SLAM system using our representation, and provide some experimental results using an RGB-D dataset.

\end{abstract}


\section{INTRODUCTION}

In simultaneous localisation and mapping (SLAM), an autonomous agent in an unknown environment needs to localise itself against the current map of its surroundings while simultaneously updating that map based on the observations it makes with its sensors.
While the choice of representation for the robot state is largely straightforward, there are many possibilities for how to represent the map.

Typically, in sparse landmark-based systems, maps are represented as sparse collections of 3D keypoints.
While state-of-the-art keypoint-based methods are very efficient and capable of highly accurate and robust pose tracking (\cite{Mur-Artal:etal:TRO2017, Engel:etal:PAMI2017}), the maps they create lack any higher-level abstractions.
For certain tasks, such as manipulation, augmented reality, or safe robotic navigation, these maps may not be sufficient.

\begin{figure}[t!]
  \centering
  \includegraphics[width=1\linewidth]{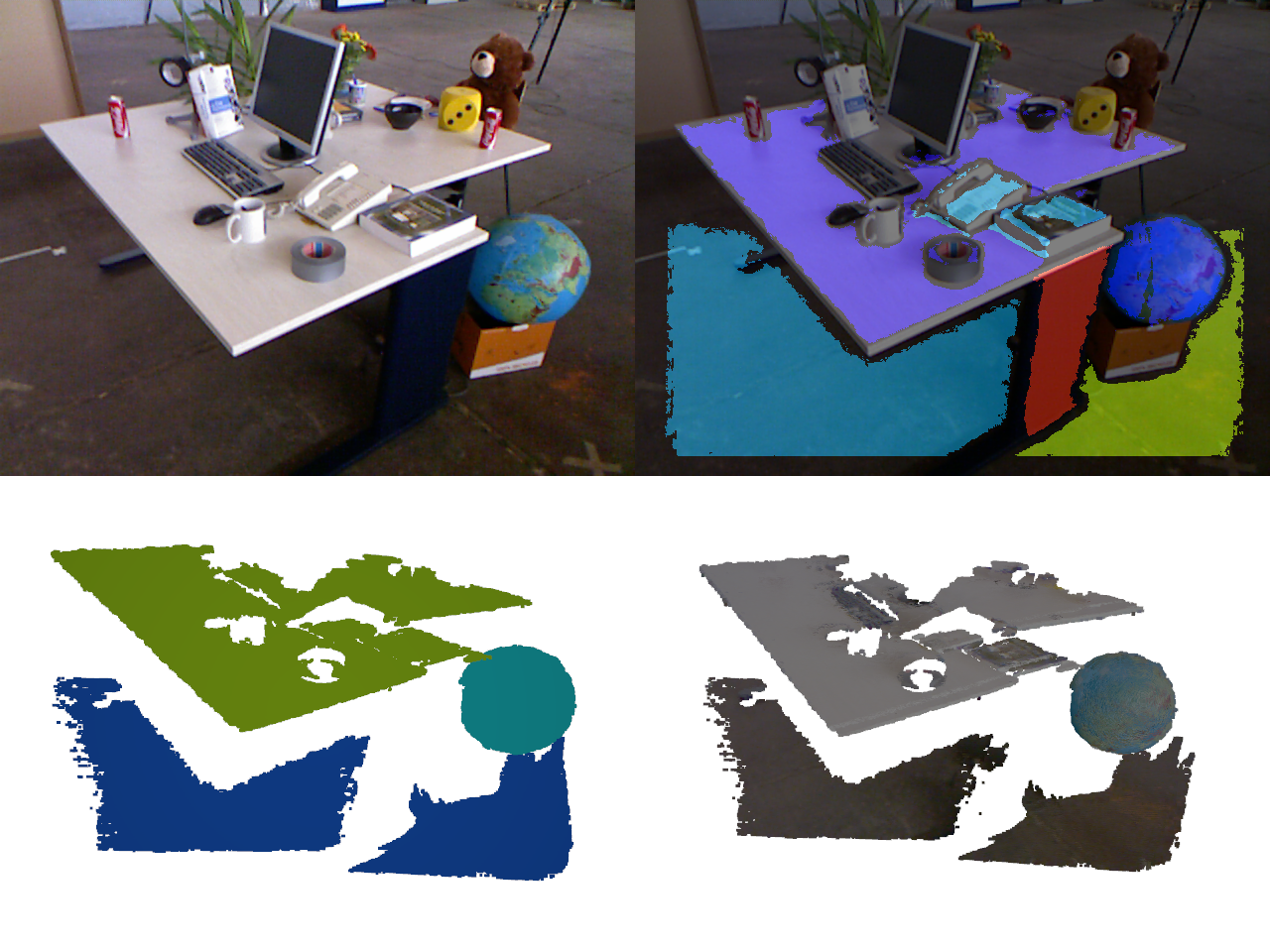}
  \caption{Example output from our proof-of-concept SLAM system using quadric surfaces as landmarks when running on the \textit{f2\_desk} sequence in the TUM RGB-D dataset \cite{Sturm:etal:IROS2012}. Top Left: Colour image. Top Right: Single-image quadric segmentation. Bottom Left: Sliding-window reconstruction of the quadrics in the map created by projecting the points associated with each quadric onto the estimated quadric surface (each quadric is assigned a unique random colour). Bottom Right: The reconstruction where each point is assigned the corresponding colour from the colour image.}
  \label{fig:teaser}
  \vspace{2mm}\hrule
  \vspace{-5mm}
\end{figure}

To address this, dense mapping systems such as DTAM \cite{Newcombe:etal:ICCV2011} and REMODE \cite{Pizzoli:etal:ICRA2014} aim to create dense reconstructions of the scene.
The advent of commodity depth cameras led to even more accurate and robust dense mapping systems such as KinectFusion \cite{Newcombe:etal:ISMAR2011}, ElasticFusion \cite{Whelan:etal:RSS2015}, and InfiniTAM \cite{Kahler:etal:ISMAR2015}.
Despite their impressive reconstructions, the resulting maps do not contain any higher-level information about the structure of the scene and often come with high demands on computation.

Human-made environments contain a lot of large-scale structure, and many SLAM systems have sought to leverage this by using higher-level features in their map representations.
Planar surfaces, in particular, have been the focus much research (e.g. \cite{Weingarten:Siegwart:IROS2005}, \cite{Salas-Moreno:etal:ISMAR2014}, \cite{Kaess:ICRA2015}, and \cite{Arndt:etal:IROS20}).
While planes are usually the dominant structure in indoor scenes, other smooth but curved structures such as cylinders and spheres are also present.
For this reason, we argue that quadric surfaces (which are also capable of representing planes) are a potentially useful landmark representation in 3D SLAM.

Almost all state-of-the-art SLAM systems are formulated as nonlinear least-squares optimisation problems.
To directly include the estimation of quadric surfaces in a least-squares optimisation requires that the representation of the quadric be minimal.
Overparameterised representations result in rank-deficient information matrices, meaning the matrix inversion necessary for using incremental solvers (such as Gauss-Newton) will fail.
While the Levenberg-Marquardt algorithm, which adds a regularisation term causing the information matrix to have full rank, can be used to avoid this problem, it is not suitable to incremental inference (\cite{Rosen:etal:2012, Rosen:etal:TRO2014}) and with an overparameterised representation, convergence will be slow.
Another possible solution is to use Lagrangian multipliers, but this comes at a cost of additional computation and a larger state space \cite{Kaess:ICRA2015}.

In this paper, we introduce a minimal representation for quadric surfaces, by decomposing the quadric representation into a 3D pose and 3D scaling parameters.
Since it is a minimal representation, it is suitable for use with Gauss-Newton, Powell's Dog Leg and incremental solvers such as iSAM2 \cite{Kaess:etal:IJRR2012}.

Our contributions are as follows:
\begin{enumerate}
    \item We present a minimal representation for quadric surfaces, suitable for least-squares estimation.
    \item We show how our representation can easily be extended to include common additional constraints on quadric surfaces such as those found in quadrics of revolution.
    \item We demonstrate how our representation can be used with a simple, proof-of-concept SLAM system run on a standard RGB-D dataset (see Fig. \ref{fig:teaser}).
\end{enumerate}


\section{RELATED WORK}

There has been a lot of interest in using higher-level features in SLAM.
Some early work focused on replacing points with lines in the estimation process (\cite{Gee:Mayol:2006, Lemaire:Lacroix:ICRA2007}).
More recently, \cite{GomezOjeda:etal:TRO2019} achieved good results by using a combination of both points and lines.

Perhaps most popular has been including planar features in the SLAM estimation formulation.
One of the earliest examples of this is \cite{Weingarten:Siegwart:IROS2005}, which estimated planes explicitly using an EKF formulation.
Other examples include \cite{Gee:etal:2007, Servant:etal:ICPR2008, Salas-Moreno:etal:ISMAR2014, Concha:Civera:IROS2015}, and \cite{Hsiao:etal:ICRA2017}.
As with lines, there has also been a lot of work looking at combining planes with other features such as points (e.g. \cite{Carranza:Calway:BMVC2010, Taguchi:etal:ICRA2013, Arndt:etal:IROS20}).

Our work is particularly inspired by \cite{Kaess:ICRA2015}, which introduces a minimal representation for planes and explicitly includes these in a least-squares estimation.
This representation was achieved by mapping the homogeneous plane parameterisation to a quaternion.
We see our work as extending these ideas to quadric surfaces.

We are proposing to explicitly include quadric surfaces in the SLAM least-squares estimation formulation.
Quadric-based representations for visual features were first introduced in \cite{Cross:Zisserman:ICCV98}.
More recently, quadric-based representations have been used in ``object-aware'' and ``structure-aware'' SLAM systems.
These systems attempt to perform SLAM using landmarks at higher levels of abstraction (such as by matching observations to a predefined database of objects \cite{Salas-Moreno:etal:CVPR2013, Sucar:etal:2020}).
A quadric representation for object-aware SLAM was recently presented in \cite{Hosseinzadeh:etal:ACCV2018} and \cite{Nicholson:etal:RAL2018}.
Similar to ours, this representation decomposes the quadric into pose and scale parameters.
Their representation, however, is constrained and only capable of representing ellipsoids (\cite{Hosseinzadeh:etal:ACCV2018} models planes using a separate representation).
In addition, \cite{Nicholson:etal:RAL2018} relies on the availability of ground-truth associations between object observations.
These systems are interested in estimating the pose and rough shape of objects and so they use the dual quadric formulation (where the ellipsoid is represented as the envelope of a set of tangent planes) as a clever method for fusing bounding box detections into the map.
Similar use of the dual quadric representation for ellipsoids was also shown in \cite{Crocco:etal:CVPR2016, Rubino:etal:TPAMI2017, Gay:etal:ICCV2017, Hosseinzadeh:etal:ICRA2019} and \cite{Liao:etal:SENSORS2020}.

In this work, we present a representation that is capable of representing all quadric surfaces.
Our representation is used to estimate the actual surface rather than a bounding volume.
Additionally, we demonstrate how our representation can be easily extended to represent quadrics of revolution, and so our representation is capable of representing planes as well.

Some concurrent research \cite{Zhen:etal:CoRR2021} also describes using general quadric surfaces as features in a least-squares estimation formulation for SLAM.
Their work presents a similar decomposition of the quadric into scale and pose parameters and also proposes a RANSAC-based front-end for their SLAM system prototype.
Whereas our work suggests adaptations to ensure our representation remains minimal for quadric of revolution, they use a constraints-based approach to handle symmetric and degenerated quadrics.


\section{MAPPING WITH QUADRIC SURFACES}

In this section we will provide a brief introduction to quadric surfaces.
We will then present our minimal representation for quadrics and discuss how to formulate SLAM with quadric surfaces as a least-squares problem, including a quadric measurement model.
Finally, we will show how our representation can be adjusted to be minimal for quadrics of revolution.

We use the following notation in this work: matrices are represented by uppercase bold symbols ($\mathbf{A}$), vectors by lowercase bold symbols ($\mathbf{a}$), and scalars by lowercase italicised symbols ($a$).
Vectors or points expressed in the reference frame $A$ are denoted ${}_A\mathbf{p}$.
The homogeneous transformation matrix that transforms homogeneous points from frame $B$ to frame $A$ is written as $\mathbf{T}_{AB}$.

\subsection{State and Quadric Representation}

The state we wish to estimate consists of sensor poses ($\mathbf{x}_0$, $\mathbf{x}_1$, \ldots, $\mathbf{x}_m$) and observed quadrics ($\boldsymbol{\pi}_0$, $\boldsymbol{\pi}_1$, \ldots, $\boldsymbol{\pi}_n$).

The position and orientation of the sensor with respect to a world reference frame is given by a 4x4 transformation matrix:
\begin{equation}
    \mathbf{x} := \mathbf{T}_{WC} \in \text{SE(3)}.
\end{equation}

Quadrics are hypersurfaces defined as the zero set of a second-degree polynomial.
In projective space, this is the set of homogeneous Euclidean coordinates, $\mathbf{z}$, that solves:
\begin{equation}
    \mathbf{z}^\text{T} \mathbf{Q} \mathbf{z} = 0,
\end{equation}
\noindent
where $\mathbf{z} \in \mathbb{P}^3$, and $\mathbf{Q} \in \mathbb{R}^{4 \times 4}$ and is symmetric (see \cite{Hartley:Zisserman:Book2004} for an introduction to projective quadrics).

A quadric surface can then be defined by the 4x4 symmetric matrix $\mathbf{Q}$.
While the symmetry means that there are ten independent elements of $\mathbf{Q}$, the quadric defines a surface in $\mathbb{P}^3$ and therefore has only 9 degrees of freedom, as one degree is needed for scale.
So although the quadric could be defined by a vector in $\mathbb{R}^{10}$, this representation would not be minimal and therefore not suitable for least-squares estimation.

\subsection{Minimal Representation}

As we would like to estimate the poses and quadrics using a least-squares formulation, we require that the representation be minimal.
Since $\mathbf{Q}$ is symmetric, it can be decomposed into the form:
\begin{equation}
    \mathbf{Q} = \mathbf{T}^\text{T} \boldsymbol{\Lambda} \mathbf{T},
\end{equation}
\noindent
where $\mathbf{T} \in \mathbb{R}^{4 \times 4}$ is an orthogonal matrix, and $\boldsymbol{\Lambda} \in \mathbb{D}^4$ (the space of diagonal matrices in $\mathbb{R}^{4 \times 4}$).

Furthermore, by Sylvester's Law of Inertia, the rows of $\mathbf{T}$ can be permuted and then scaled by a diagonal matrix such that:
\begin{equation}
    \mathbf{Q} = \hat{\mathbf{T}}^\text{T} \left[ \begin{matrix} \mathbf{S} & \mathbf{0} \\ \mathbf{0}^\text{T} & 1 \end{matrix} \right]^\text{T} \mathbf{D}(\boldsymbol{\sigma}) \left[ \begin{matrix} \mathbf{S} & \mathbf{0} \\ \mathbf{0}^\text{T} & 1 \end{matrix} \right] \hat{\mathbf{T}},
\end{equation}
\noindent
where $\hat{\mathbf{T}} \in \mathbb{R}^{4 \times 4}$ is the permuted matrix $\mathbf{T}$, $\mathbf{S} \in \mathbb{D}^3$ is the scaling matrix, $\mathbf{D}(\cdot)$ is a function that returns a matrix in $\mathbb{D}^k$ with diagonal elements corresponding to the input vector in $\mathbb{R}^k$, and $\boldsymbol{\sigma} \in \mathbb{R}^4$ is the ``signature" vector and consists only of +1, -1 and 0 entries, in that order.
This decomposition is discussed in some detail in \cite{Hartley:Zisserman:Book2004}.

The signature is unique to all quadrics of a given class.
In total there are 17 different signatures, of which 9 correspond to real quadric surfaces.
For example, ${\boldsymbol{\sigma} = \left[ +1, +1, +1, -1 \right]^\text{T}}$ is shared between all ellipsoids.

We can think of $\mathbf{D}(\boldsymbol{\sigma})$ as representing a ``canonical" quadric, with unit scaling and its principal axes aligned with the world frame.
An illustration of this is given in the first row of Table \ref{tab:canonical}.
For $\boldsymbol{\sigma} = \left[ +1, +1, +1, -1 \right]^\text{T}$, if $\mathbf{T}$ and $\mathbf{S}$ are set to identity, then $\mathbf{Q}$ represents the implicit equation $x^2 + y^2 + z^2 = 1$, which is a unit sphere centred at the origin of the world frame.

\begin{table}[t!]
    \centering
    \begin{tabular}{c c c}
         \toprule
         $\mathbf{T}_{WQ}$ & $\mathbf{S}$ & \textbf{Quadric} \\
         \midrule
         $\left[ \begin{matrix} 1 & 0 & 0 & 0 \\
                                0 & 1 & 0 & 0 \\
                                0 & 0 & 1 & 0 \\
                                0 & 0 & 0 & 1 \end{matrix} \right]$ &
         $\left[ \begin{matrix} 1 & 0 & 0 \\
                                0 & 1 & 0 \\
                                0 & 0 & 1 \end{matrix} \right]$ &
         \raisebox{-0.5\totalheight}{\includegraphics[width=0.2\linewidth]{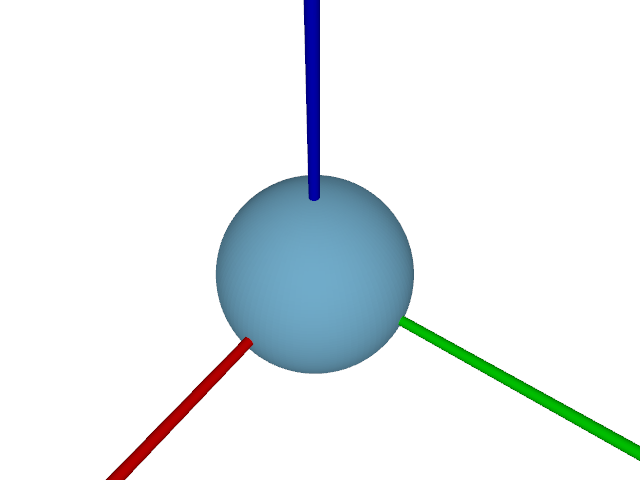}} \\
         \midrule
         $\left[ \begin{matrix} 1 & 0 & 0 & 0 \\
                                0 & 1 & 0 & 0 \\
                                0 & 0 & 1 & 0 \\
                                0 & 0 & 0 & 1 \end{matrix} \right]$ &
         $\left[ \begin{matrix} 1 & 0 & 0 \\
                                0 & 0.5 & 0 \\
                                0 & 0 & 2 \end{matrix} \right]$ &
         \raisebox{-0.5\totalheight}{\includegraphics[width=0.2\linewidth]{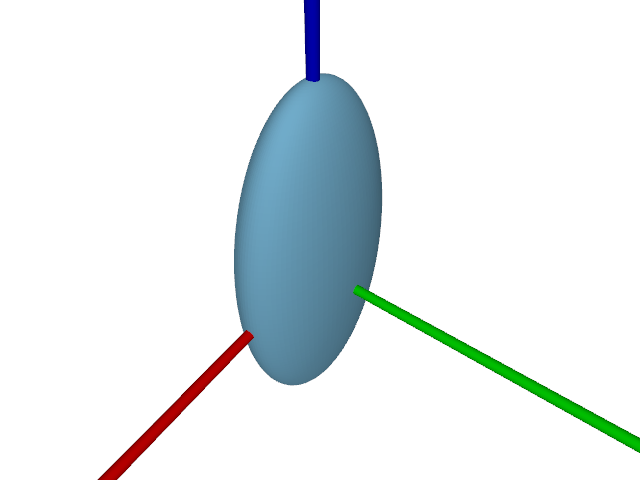}} \\
         \midrule
         $\left[ \begin{matrix} 1 & 0 & 0 & 0 \\
                                0 & 0 & 1 & 3 \\
                                0 & -1 & 0 & 0 \\
                                0 & 0 & 0 & 1 \end{matrix} \right]$ &
         $\left[ \begin{matrix} 1 & 0 & 0 \\
                                0 & 0.5 & 0 \\
                                0 & 0 & 2 \end{matrix} \right]$ &
         \raisebox{-0.5\totalheight}{\includegraphics[width=0.2\linewidth]{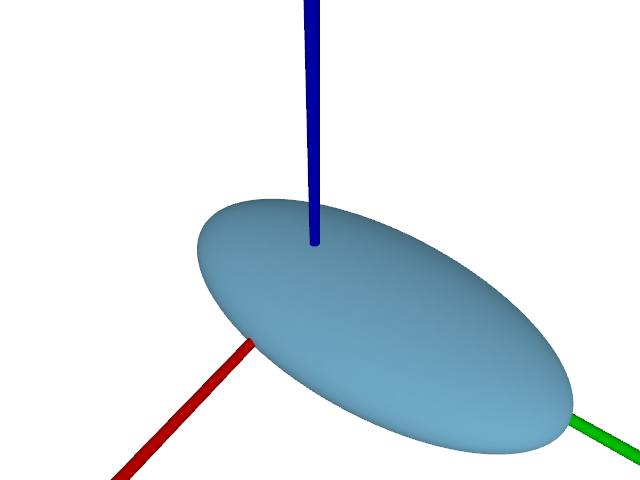}} \\
         \bottomrule
    \end{tabular}
    \caption{A demonstration of how the diagonal matrix, $\mathbf{D}(\boldsymbol{\sigma})$, represents a canonical quadric that is scaled and transformed by the $\mathbf{S}$ and $\mathbf{T}_{WQ}$ matrices, respectively. For $\boldsymbol{\sigma} = \left[ +1, +1, +1, -1 \right]^\text{T}$, when $\mathbf{T}_{WQ}$ and $\mathbf{S}$ are both identity, $\mathbf{Q}$ represents a unit sphere centred at the origin of the world frame. Changing the diagonal entries of $\mathbf{S}$ scales the quadric to an ellipsoid, and changing $\mathbf{T}_{WQ}$ changes its pose.}
    \label{tab:canonical}
    \vspace{2mm}\hrule
    \vspace{-5mm}
\end{table}

By defining $\mathbf{S}$ as:
\begin{equation}
    \mathbf{S} = \left[ \begin{matrix} \alpha & 0     & 0 \\
                                     0 & \beta & 0 \\
                                     0 & 0     & \gamma \end{matrix} \right],
\end{equation}
\noindent
and $\hat{\mathbf{T}}$ as the homogeneous transformation matrix $\mathbf{T}_{WQ}$ (which maps from the canonical quadric frame to the world frame), it is equivalent to first scaling the canonical quadric along each of its principal axes and then transforming it to a desired position and orientation.
This is shown for two examples in Table \ref{tab:canonical}.

We can then fully define a quadric by this decomposition:
\begin{equation}
    \boldsymbol{\pi} := \left( \mathbf{T}_{WQ}, \: \mathbf{S}, \: \boldsymbol{\sigma} \right) \in \text{SE(3)} \times \mathbb{D}^3 \times \mathbb{R}^4.
\end{equation}

Assuming that the class of quadric surface is determined by the front-end of the SLAM system (and that the signature is therefore a known quantity), the parameters to estimate can be simplified to:
\begin{equation}
    \boldsymbol{\pi}^{(\boldsymbol{\sigma})} := \left( \mathbf{T}_{WQ}, \: \mathbf{S} \right) \in \text{SE(3)} \times \mathbb{D}^3.
\end{equation}
\noindent
As we will use this assumption throughout this work, we will drop the superscript going forward.

The minimal representation for the quadric is then given by:
\begin{equation}
    \delta\boldsymbol{\pi} = \left[ \boldsymbol{\xi}^\text{T}, \: \mathbf{s}^\text{T} \right]^\text{T} \in \mathbb{R}^9,
\end{equation}
\noindent
where $\boldsymbol{\xi} \in \mathbb{R}^6$ is the twist representing the quadric pose and $\mathbf{s} \in \mathbb{R}^3$ represents the scale parameters.

To map between the state and minimal representations, we define a compound $\boxplus$-manifold \cite{Hertzberg:etal:INFORMATIONFUSION2011}, consisting of $\boxplus$-manifolds over $\text{SE(3)}$ and $\mathbb{D}^3$:
\begin{align}
    \boldsymbol{\pi} &\boxplus \delta\boldsymbol{\pi} := (\mathbf{T}_{WQ} \boxplus_\text{SE(3)} \boldsymbol{\xi}, \ \mathbf{S} \boxplus_{\mathbb{D}^3} \mathbf{s} ) \\
    \boldsymbol{\pi} &\boxminus \boldsymbol{\pi}' := (\mathbf{T}_{WQ} \boxminus_\text{SE(3)} \mathbf{T}_{WQ}', \ \mathbf{S} \boxminus_{\mathbb{D}^3} \mathbf{S}').
\end{align}
\noindent
The $\boxplus$-manifold over $\text{SE(3)}$ is given by:
\begin{equation}
    \mathbf{T}_{WQ} \boxplus_{\text{SE(3)}} \boldsymbol{\xi} = \exp{(\boldsymbol{\xi}^\wedge)} \mathbf{T}_{WQ}
\end{equation}
\noindent
where
\begin{align}
    \boldsymbol{\xi}^\wedge &= \left[ \begin{matrix} \boldsymbol{\rho} \\ \boldsymbol{\phi} \end{matrix} \right] = \left[ \begin{matrix} \boldsymbol{\phi}^\times & \boldsymbol{\rho} \\ \mathbf{0}^\text{T} & 0 \end{matrix} \right] \in \mathbb{R}^{4 \times 4}, \\
    \boldsymbol{\phi}^\times &= \left[ \begin{matrix} 0 & -\phi_3 & \phi_2 \\ \phi_3 & 0 & -\phi_1 \\ -\phi_2 & \phi_1 & 0 \end{matrix} \right] \in \mathbb{R}^{3 \times 3},
\end{align}
\noindent
$\boldsymbol{\rho} \in \mathbb{R}^3$ is the minimal parameterisation of the quadric position, $\boldsymbol{\phi} \in \mathbb{R}^3$ is the minimal parameterisation of the quadric orientation, and
\begin{equation}
    \mathbf{T}_{WQ} \boxminus_{\text{SE(3)}} \mathbf{T}_{WQ}' = \log{({\mathbf{T'}_{WQ}^{-1}} \mathbf{T}_{WQ})}^\vee
\end{equation}
\noindent
where
\begin{equation}
    \left[ \begin{matrix} \boldsymbol{\phi}^\times & \boldsymbol{\rho} \\ \mathbf{0}^\text{T} & 0 \end{matrix} \right]^\vee = \left[ \begin{matrix} \boldsymbol{\rho} \\ \boldsymbol{\phi} \end{matrix} \right].
\end{equation}

The $\boxplus$-manifold over $\mathbb{D}^3$ is given by:
\begin{align}
    \mathbf{S} &\boxplus_{\mathbb{D}^3} \mathbf{s} = \mathbf{S} + \mathbf{D}(\mathbf{s}) \\ 
    \mathbf{S} &\boxminus_{\mathbb{D}^3} \mathbf{S}' = \mathbf{D}^{-1}(\mathbf{S} - \mathbf{S}'),
\end{align}
\noindent
where $\mathbf{D}^{-1}(\cdot)$ is the inverse of the function $\mathbf{D}(\cdot)$, mapping the diagonal elements of the input matrix in $\mathbb{D}^k$ to a vector in $\mathbb{R}^k$.

\subsection{SLAM Formulation}

\begin{figure}[t!]
  \centering
  \includegraphics[trim=80 120 70 60,clip,width=0.9\linewidth]{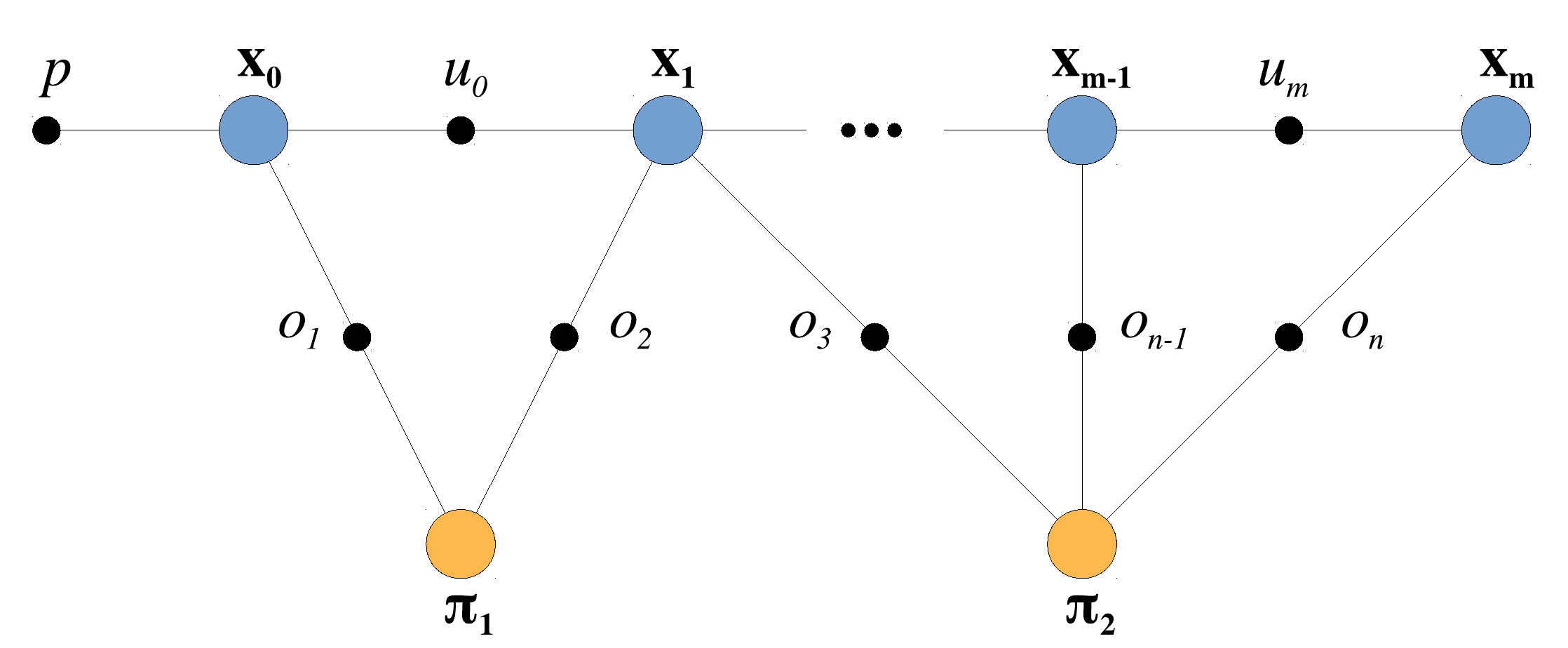}
  \caption{The factor graph for the SLAM formulation with quadric surfaces. Variable nodes in the graph represent sensor poses (\textbf{x}) and quadrics ($\boldsymbol{\pi}$). Factor nodes in the graph represent odometry measurements (\textit{u}), quadric observations (\textit{o}), and a prior on the first pose (\textit{p}).}
  \label{fig:factor_graph}
  \vspace{2mm}\hrule
  \vspace{-5mm}
\end{figure}

We formulate the quadric mapping problem as a least-squares optimisation, using a factor graph to represent the estimation problem as a graphical model \cite{Kaess:etal:IJRR2012}.
The factor graph of our problem is shown in Fig.~\ref{fig:factor_graph}.

SLAM using quadrics is very similar to sparse keypoint-based SLAM.
We want to jointly estimate the pose of the sensor in each frame ($\mathbf{x}_0$, $\mathbf{x}_1$, \ldots, $\mathbf{x}_m$) along with the parameters of all observed quadric surfaces ($\boldsymbol{\pi}_0$, $\boldsymbol{\pi}_1$, \ldots, $\boldsymbol{\pi}_n$).

\subsection{Quadric Measurement Model}

We model the uncertainty in a measurement of a quadric surface  ${}_C\boldsymbol{\pi}$ from pose $\mathbf{T}_{WC}$ by zero-mean Gaussian noise $\boldsymbol{\nu}$ with a 9x9 covariance matrix $\boldsymbol{\Sigma}$:
\begin{equation}
    {}_C\boldsymbol{\pi} = h(\mathbf{T}_{WC}, \boldsymbol{\pi}) \boxplus_\text{SE(3)} \boldsymbol{\nu}, \quad \boldsymbol{\nu} \sim \mathcal{N}(0, \boldsymbol{\Sigma}),
\end{equation}
\noindent
where $h(\cdot, \cdot)$ is the measurement prediction function that transforms the quadric from the world frame to the sensor frame and leaves the scale parameters untouched:
\begin{equation}
    h(\mathbf{T}_{WC}, \boldsymbol{\pi}) = (\mathbf{T}_{WC}^{-1} \mathbf{T}_{WQ}, \: \mathbf{S}).
\end{equation}

The probability of a quadric estimate $\hat{\boldsymbol{\pi}}$ and pose estimate $\hat{\mathbf{x}}$ given a measurement ${}_C\Tilde{\boldsymbol{\pi}}$ is given by:
\begin{align}
    p(\hat{\mathbf{x}}, \hat{\boldsymbol{\pi}} | {}_C\Tilde{\boldsymbol{\pi}} ) &= \eta \exp{\left\{ -\frac{1}{2} \lVert h(\mathbf{T}_{W\hat{C}}, \hat{\boldsymbol{\pi}}) \boxminus {}_C\Tilde{\boldsymbol{\pi}} \rVert_{\boldsymbol{\Sigma}}^2 \right\} }, \\
    \eta &= \frac{1}{\sqrt{(2 \pi)^9 | \boldsymbol{\Sigma} |}}.
\end{align}

Therefore, the least-squares cost function that minimises the negative log-likelihood is given by:
\begin{equation}
    c(\hat{\mathbf{x}}, \hat{\boldsymbol{\pi}}) = \frac{1}{2} \lVert h(\hat{\mathbf{T}}_{WC}, \boldsymbol{\pi}) \boxminus {}_C\Tilde{\boldsymbol{\pi}} \rVert_{\boldsymbol{\Sigma}}^2.
    \label{eq:abs}
\end{equation}
\noindent
This is the cost function used for the observation factors in the factor graph (see Figure \ref{fig:factor_graph}).

\subsection{Quadrics of Revolution}

\begin{table*}[t!]
    \centering
    \begin{tabular}{c c c c c c c}
         \toprule
         \textbf{Quadric} & \textbf{Signature} & \textbf{Minimal Parameterisation} & \textbf{Position} & \textbf{Orientation} & \textbf{Scale} & $\partial \boldsymbol{\xi} / \partial \boldsymbol{\xi}_\text{Q}$ \\
         \midrule
         General & -- & ($\rho_1$, $\rho_2$, $\rho_3$, $\phi_1$, $\phi_2$, $\phi_3$, $\alpha$, $\beta$, $\gamma$) & ($\rho_1$, $\rho_2$, $\rho_3$) & ($\phi_1$, $\phi_2$, $\phi_3$)  & ($\alpha$, $\beta$, $\gamma$) & -- \\
         \midrule
         Plane & (+1, 0, 0, 0) & ($\rho_1$, $\phi_2$, $\phi_3$) & ($\rho_1$, $0$, $0$) & ($0$, $\phi_2$, $\phi_3$) & ($1$, $1$, $1$) & $\left[ \begin{matrix} 1 & 0 & 0 & 0 & 0 & 0 \\ 0 & 0 & 0 & 0 & 1 & 0 \\ 0 & 0 & 0 & 0 & 0 & 1 \end{matrix} \right]^\text{T}$ \\
         \midrule
         Sphere & (+1, +1, +1, -1) & ($\rho_1$, $\rho_2$, $\rho_3$, $\alpha$) & ($\rho_1$, $\rho_2$, $\rho_3$) & ($0$, $0$, $0$) & ($\alpha$, $\alpha$, $\alpha$) & $\left[ \begin{matrix} 1 & 0 & 0 & 0 & 0 & 0 \\ 0 & 1 & 0 & 0 & 0 & 0 \\ 0 & 0 & 1 & 0 & 0 & 0 \end{matrix} \right]^\text{T}$ \\
         \midrule
         \makecell{Circular \\ Cylinder} & (+1, +1, 0, -1) & ($\rho_1$, $\rho_2$, $\phi_1$, $\phi_2$, $\alpha$) & ($\rho_1$, $\rho_2$, $0$) & ($\phi_1$, $\phi_2$, $0$) & ($\alpha$, $\alpha$, $1$) & $\left[ \begin{matrix} 1 & 0 & 0 & 1 & 0 & 0 \\ 0 & 1 & 0 & 0 & 1 & 0 \\ 0 & 0 & 0 & 0 & 0 & 0 \end{matrix} \right]^\text{T}$ \\
         \midrule
         \makecell{Circular \\ Cone} & (+1, +1, -1, 0) & ($\rho_1$, $\rho_2$, $\rho_3$, $\phi_1$, $\phi_2$, $\alpha$) & ($\rho_1$, $\rho_2$, $\rho_3$) & ($\phi_1$, $\phi_2$, $0$) & ($\alpha$, $\alpha$, $1$) & $\left[ \begin{matrix} 1 & 0 & 0 & 1 & 0 & 0 \\ 0 & 1 & 0 & 0 & 1 & 0 \\ 0 & 0 & 1 & 0 & 0 & 0 \end{matrix} \right]^\text{T}$ \\
         \bottomrule
    \end{tabular}
    \caption{Our representation for general quadric surfaces can be easily extended to provide minimal representations for quadrics of revolution. Here, examples are provided by planes, spheres, circular cylinders, and circular cones.}
    \label{tab:quadrics}
    \vspace{2mm}\hrule
\end{table*}

While general quadrics are a potentially useful representation for SLAM, there are a number of additional constraints that we may wish to place on quadric surfaces.
For example, quadrics of revolution (such as spheroids, paraboloids, circular cones, and circular cylinders) are quadrics where some or all of the scale parameters are constrained to be equal.
These constrained quadrics are of particular interest, as they occur regularly in human-made environments and there are many off-the-shelf RANSAC-based methods designed to quickly detect these types of surfaces from point clouds (e.g. \cite{Schnabel:etal:CGF2007}).

Although our proposed representation is minimal for general 2D quadric surfaces, it is \textit{not} minimal for these constrained quadrics.
This means that using our general quadric representation for quadrics of revolution will result in rank-deficient information matrices, which is not suitable for least-squares estimation.
To see why, consider the sphere in Table \ref{tab:canonical}.
For spheres, all scale parameters are constrained to be equal, meaning that the surface is invariant to all rotations.
Similarly, circular cylinders and cones are invariant to rotations about their axes.
Some quadrics can also be invariant to certain translations.
For example, infinite planes are invariant to translations perpendicular to their surface normal.
Infinite planes are also invariant to the scaling parameters.

Our representation can be easily adapted to handle these constrained quadrics by reducing the minimal parameter space to exclude the invariant transformations and scaling.

The Jacobians of these constrained quadrics are also simple adaptations of the Jacobian for the general quadric.
For example, since the sphere is invariant to rotation, instead of estimating ${\boldsymbol{\xi} = \left[ \boldsymbol{\rho}^\text{T}, \boldsymbol{\phi}^\text{T} \right]^\text{T}}$ where $\boldsymbol{\rho} \in \mathbb{R}^3$ is the minimal parameterisation for the quadric position and $\boldsymbol{\phi} \in \mathbb{R}^3$ is the minimal parameterisation for the quadric orientation, we estimate ${\boldsymbol{\xi}_\text{sphere} = \boldsymbol{\rho}}$.
The derivative of the residual with respect to the optimisation parameters is now $\partial e/\partial \boldsymbol{\xi}_\text{sphere}$ rather than $\partial e/\partial \boldsymbol{\xi}$, but this can be easily amended by chaining a simple matrix to the Jacobian for the general quadric:
\begin{align}
    \frac{\partial e}{\partial \boldsymbol{\xi}_\text{sphere}} &= \frac{\partial e}{\partial \boldsymbol{\xi}} \frac{\partial \boldsymbol{\xi}}{\partial \boldsymbol{\xi}_\text{sphere}} \\
    &= \frac{\partial e}{\partial \boldsymbol{\xi}} \left[ \begin{matrix} \mathbf{I}_{3 \times 3} \\ \mathbf{0}_{3 \times 3} \end{matrix} \right].
\end{align}

\begin{figure*}[t!]
  \centering
  \includegraphics[width=0.9\textwidth]{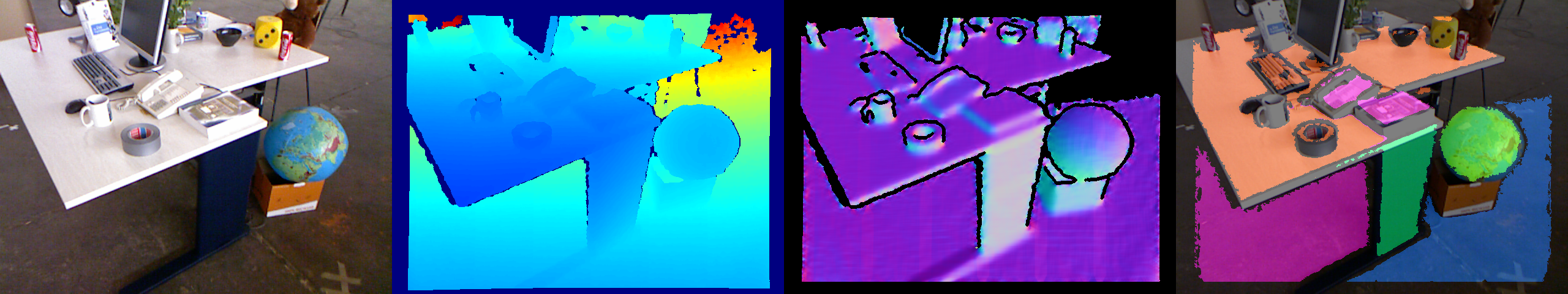} \\
  \vspace{0.5mm}
  \includegraphics[width=0.9\textwidth]{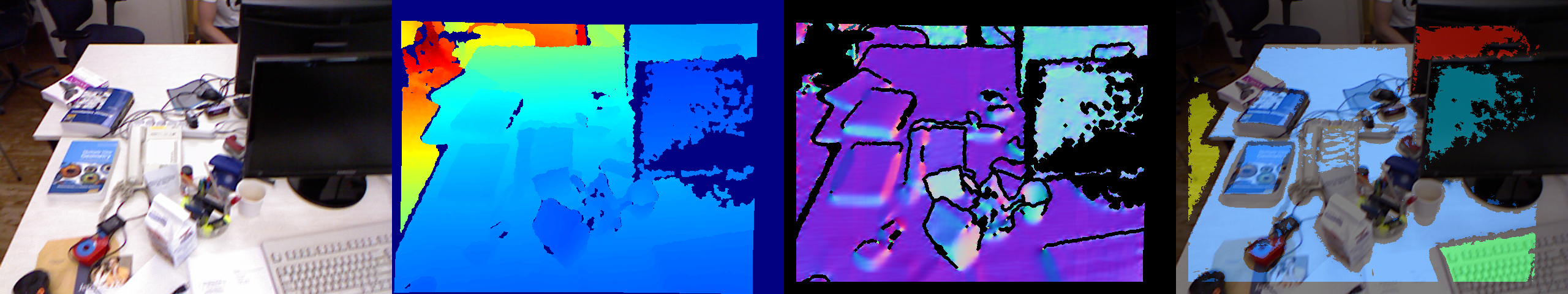}
  \caption{Our proof-of-concept quadric SLAM system takes depth maps as input. For each depth map, we estimate the surface normals and then use the backprojected depth points and normals to segment the input into various quadrics. From left to right: colour, depth, normals, and quadric segmentation for a single frame in the TUM RGB-D dataset \cite{Sturm:etal:IROS2012}.}
  \label{fig:normals_and_seg}
  \vspace{2mm}\hrule
  \vspace{-5mm}
\end{figure*}

In Table \ref{tab:quadrics}, we provide several adaptations for common constrained quadrics.
Note that for the circular cylinder, we have changed the order of the signature for simplicity.
This is because the cylinder is a cone with its apex at infinity and so this permutation of rows is required.
This permutation is normally covered by the orthonormal matrix, but to have a canonical right cylinder in the quadric frame, we start with a permuted signature.


\section{EXPERIMENTAL RESULTS}

To demonstrate how our representation can be used, we introduce a proof-of-concept SLAM system using quadrics as landmarks and present some results from the TUM RGB-D dataset \cite{Sturm:etal:IROS2012}.
All experiments are run on a laptop computer with an i7-6700HQ 2.60GHz CPU.

Our system takes depth maps as input.
For each depth map, we backproject the points to 3D and estimate the corresponding surface normals for the point cloud using the PCL 1.11 implementation of \cite{Holzer:etal:IROS2012}.

Using the points and normals, we estimate the frame-to-frame odometry using an ICP point-to-plane technique.

We then use the normals to help segment the point cloud into planes, spheres, circular cylinders and circular cones using the CGAL 5.2 implementation of Efficient RANSAC \cite{Schnabel:etal:CGF2007}.
The Efficient RANSAC algorithm samples a set of points, fits candidate shapes to the sampled set, and then evaluates how well the remaining points are approximated by the candidate shape.
After a number of attempts, the shape that fit the most points is extracted, and the algorithm continues by sampling from the remaining points.
While the CGAL 5.2 implementation does not output the resulting quadrics in our representation, the conversion to our representation is trivial.
For example, when fitting a sphere, Efficient RANSAC gives the centre and radius, which are simply the translation component of $\mathbf{T}_{WQ}$ and the inverse of the scale parameter, respectively.
An example of the estimated normals and resulting segmentation is presented in Fig. \ref{fig:normals_and_seg}.

The observed quadrics are then matched to those quadrics currently in the map.
To do the data association, we simply compare each observed quadric against all other quadrics of the same class in the map and match it with the quadric that has the lowest error under a threshold.
The matching error is calculated using the cost function in (\ref{eq:abs}), where the estimated sensor pose is the estimated previous pose concatenated with the latest odometry measurement.

The new pose and any new quadrics are added to the factor graph, along with the odometry and quadric measurements.
To prevent any spurious quadric detections from polluting the map, we require a quadric to have been observed in 5 frames before it is added.
The graph is then optimised using an iterative dog leg method \cite{Rosen:etal:TRO2014} in GTSAM 4.0 \cite{Dellaert:TechReport2012}.
Note that we are not doing any marginalisation and so are performing an entire batch optimisation with each step.

Results from the \textit{fr2\_desk} sequence of the TUM RGB-D dataset \cite{Sturm:etal:IROS2012} are shown in Fig.~\ref{fig:teaser}.
The front-end of the system takes the depth map as input (not shown) and outputs the quadric segmentation shown in the top right.
In this case, the front-end has detected 5 planes (two on the floor, the top of the desk, the side of the desk, and a small region corresponding with the phone and textbook), and a sphere (the globe).
These are then matched with quadrics already in the map leading to matches for both floor planes, the top of the desk and the globe.
A sliding-window reconstruction of the map at this stage is shown in the bottom left where each quadric has been assigned a random colour (both floor planes have been matched with the same quadric).

\begin{figure}
    \centering
    \includegraphics[width=1\linewidth]{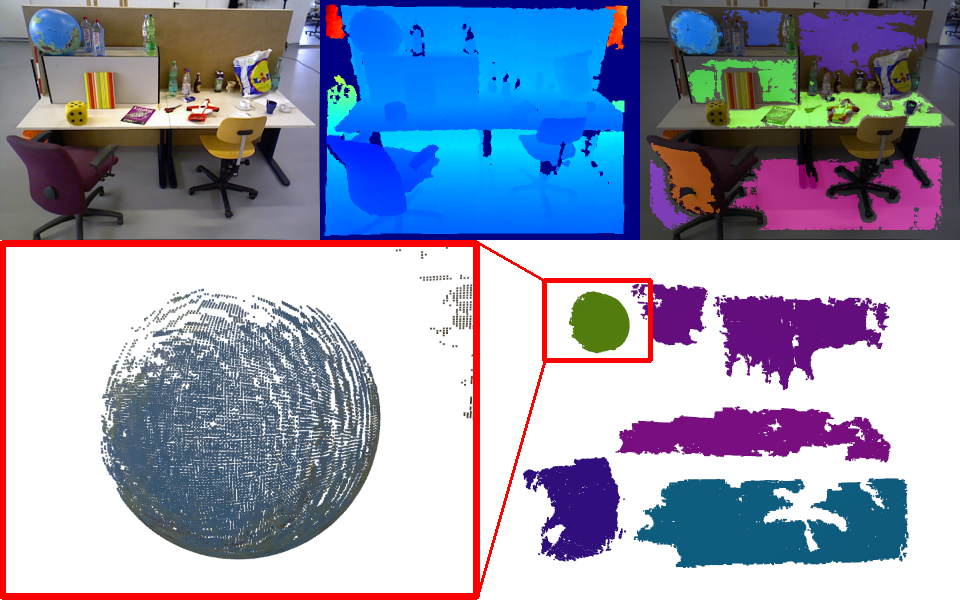}
    \caption{Example output when running on the \textit{f3\_long\_office\_household} sequence from the TUM RGB-D dataset \cite{Sturm:etal:IROS2012}. Top Row: Colour, depth, and quadric segmentation from a single frame. Bottom Left: Close up on globe (notice it is a perfect sphere). Bottom Right: A sliding window reconstruction of the quadrics in the map created by projecting the points associated with each quadric onto the estimated quadric surface (each quadric is assigned a unique random colour). }
    \label{fig:results}
    \vspace{2mm}\hrule
    \vspace{-5mm}
\end{figure}

Results from the \textit{fr3\_long\_office\_household} sequence are shown in Fig.~\ref{fig:results}.
The top row shows the input colour and depth images, along with the detected quadrics.
The bottom right shows a sliding window reconstruction of the point cloud projected onto the estimated quadrics, with each unique quadric being assigned a random colour.
The bottom left shows a close up of the reconstruction of the globe where each point is assigned the corresponding colour from the image.
Since our representation enforces constraints on the quadric, the resulting reconstruction is a perfect sphere.

We present runtime statistics for key components of our system in Table \ref{tab:timing}.
As currently implemented, the front-end of our system is capable of running at approximately 5 Hz.
As we are currently doing a full batch optimisation with each frame, the time required to optimise the graph grows as the graph increases in size.
Using more sophisticated techniques such as bounded window keyframing with marginalisation or iterative methods such as iSAM2 \cite{Kaess:etal:IJRR2012} should speed up the per frame optimisation time considerably, particularly during long sequences.

\begin{table}[]
    \centering
    \begin{tabular}{c c}
         \toprule
         \textbf{System Component} & \textbf{Median Runtime (ms)} \\
         \midrule
         Normal Computation & 18 \\
         \midrule
         Quadric Segmentation & 156 \\
         \midrule
         Data Association & 5 \\
         \midrule
         Graph Optimisation & 12 \\
         \bottomrule
    \end{tabular}
    \caption{The per frame runtime for key components in our proof-of-concept SLAM system. Note that since we do a full batch optimisation for each frame, the optimisation time grows as the graph increases in size. Using more sophisticated techniques such as bounded window keyframing with marginalisation or iterative methods such as iSAM2 \cite{Kaess:etal:IJRR2012} would be obvious improvements.}
    \label{tab:timing}
    \vspace{2mm}\hrule
    \vspace{-5mm}
\end{table}


\section{CONCLUSION}

As human-made environments typically have a lot of structure (planes, but also other smooth structures such as cylinders and spheres), we believe that quadric surfaces are a useful representation in 3D SLAM.
To this end, we have introduced a minimal representation for quadric surfaces, suitable for inclusion in least-squares estimation formulations.
Our representation is based on a decomposition of the homogeneous quadric representation into pose and scale parameters.
Unlike most existing quadric representations in SLAM, our representation is capable of handling all quadric surfaces, not just ellipsoids.
As a proof-of-concept, we have presented a simple SLAM system using our representation.

As this work has demonstrated that quadrics can be included in a least-squares formulation, future work can focus on developing more sophisticated 3D SLAM systems using quadric surfaces.
To make an accurate and robust real-time 3D SLAM system using our representation, several improvements would need to be made.
The quadric segmentation is the current bottleneck in the front-end of our system and prevents it from running at more than 5 Hz.
One possible solution is to include other features, such as points, in the system, and to do frame-to-frame tracking with the simpler representations, detecting quadrics only on more infrequent keyframes.
Including these additional features would also help in situations where there are insufficient constraints on the quadrics due to a lack of observations.
Another possible speed up is to try and predict the presence of quadrics using a trained neural network, amortising the search cost.
Performance improvements could also be made on the back-end of our system by replacing the batch optimisation on every frame with an iterative solver such as iSAM2 \cite{Kaess:etal:IJRR2012} or using bounded window keyframing with marginalisation.

While our method uses a quadric representation to estimate actual surfaces rather than rough bounding volumes as in \cite{Hosseinzadeh:etal:ACCV2018} and \cite{Nicholson:etal:RAL2018}, it assumes that the quadric parameters are directly observable.
Although we have shown that this can be done by using off-the-shelf algorithms such as Efficient RANSAC \cite{Schnabel:etal:CGF2007}, it is slower and possibly less robust than the implicit quadric detections used in those other methods.
An interesting possible future direction would be the combination of these two approaches: using the rough bounding volumes to detect objects and split the point cloud, and using our proposed method to estimate surfaces within those volumes.


\addtolength{\textheight}{-2.5cm}   

\bibliographystyle{IEEEtran}
\bibliography{robotvision}


\end{document}